\pdfoutput=1

\documentclass[11pt]{article}

\usepackage[]{naacl2021}

\usepackage{times}
\usepackage{latexsym}

\usepackage[T1]{fontenc}

\usepackage[utf8]{inputenc}

\usepackage{microtype}
\usepackage{graphicx}
\usepackage{amsmath}
\usepackage{amsthm}
\usepackage{amssymb}
\usepackage{booktabs}
\usepackage{tabularx}
\usepackage{algorithm}
\usepackage{algorithmic}
\usepackage{subfigure}
\usepackage{paralist}
\usepackage{amssymb}
\usepackage{bm}
\usepackage{color}
\definecolor{newblue}{RGB}{70,118,214}
\definecolor{neworange}{RGB}{206,151,90}
\definecolor{newpurple}{RGB}{131,106,212}
\definecolor{newgreen}{RGB}{58,105,57}
\title{Alleviating the Knowledge-Language Inconsistency: A Study for \\Deep Commonsense Knowledge}

  \author{Yi Zhang\textsuperscript{$\dag$}\footnotemark[1], Lei Li\textsuperscript{$\dag$}\Thanks{\ Equal contribution}, Yunfang Wu\textsuperscript{$\dag$}, Qi Su\textsuperscript{$\dag\S$}, Xu Sun\textsuperscript{$\dag$}\\
   \textsuperscript{$\dag$} MOE Key Laboratory of Computational Linguistics, School of EECS, Peking University \\

  \textsuperscript{$\S$} School of Foreign Languages, Peking University\\
    \texttt{\{zhangyi16,wuyf,sukia,xusun\}@pku.edu.cn}\\
    \texttt{lilei@stu.pku.edu.cn} \\
  }

\begin{document}
\maketitle
\begin{abstract}
Knowledge facts are typically represented by relational triples, while we observe that some commonsense facts are represented by the triples whose forms are inconsistent with the expression of language. This inconsistency puts forward a challenge for pre-trained language models to deal with these commonsense knowledge facts.
In this paper, we term such knowledge as \emph{deep commonsense knowledge} and conduct extensive exploratory experiments on it. We show that deep commonsense knowledge occupies a significant part of commonsense knowledge while conventional methods fail to capture it effectively. We further propose a novel method to mine the deep commonsense knowledge distributed in sentences, alleviating the reliance of conventional methods on the triple representation form of knowledge. Experiments demonstrate that the proposal significantly improves the performance in mining deep commonsense knowledge.
\end{abstract}

\section{Introduction}

\begin{figure}
    \centering
    \includegraphics[width=0.9\linewidth]{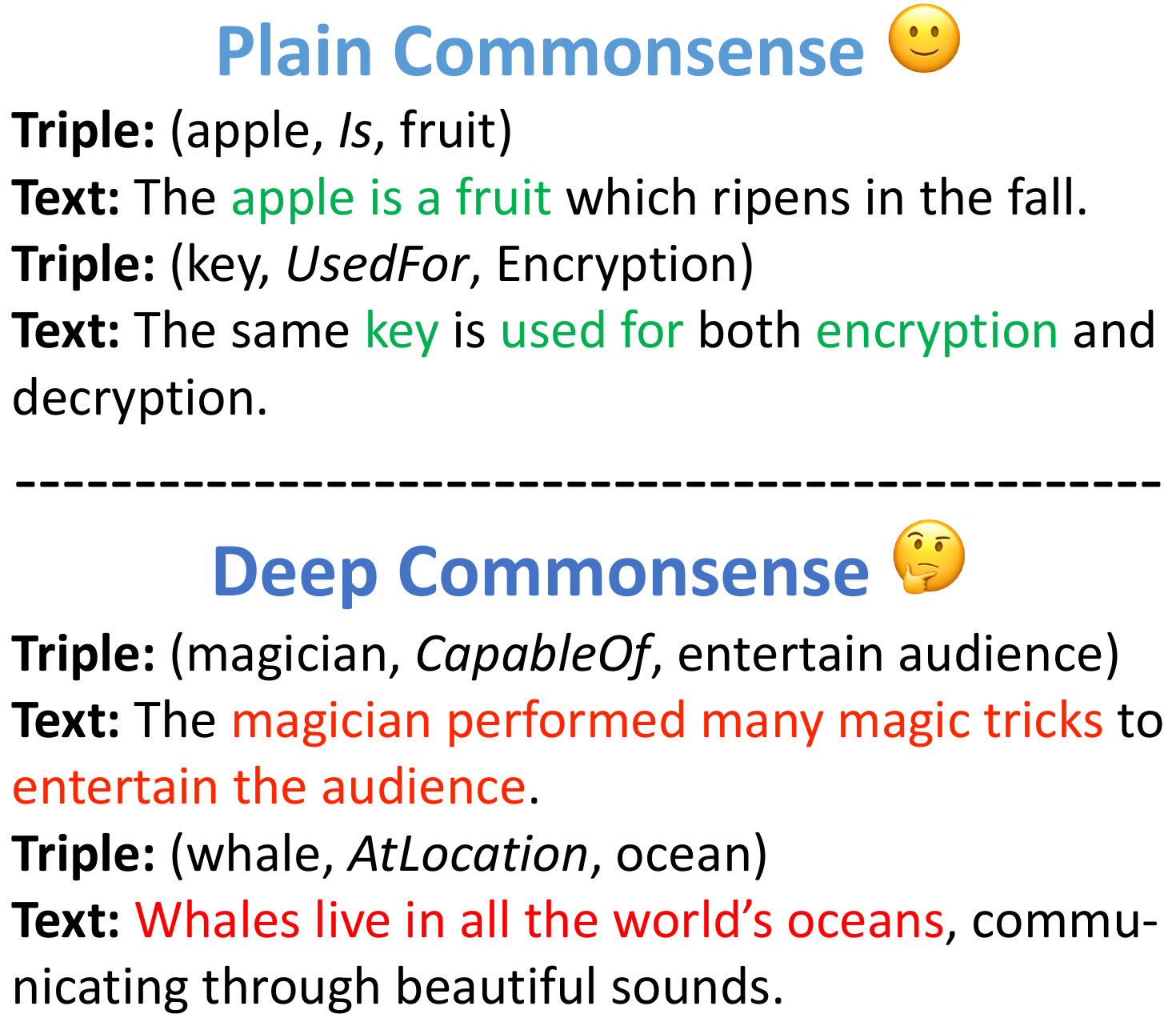}
    \caption{Typical examples of plain commonsense and deep commonsense knowledge. “Text” denotes the corresponding language where the knowledge triple lies behind.}
    \label{fig:examples}
\end{figure}

The typical representation of a commonsense knowledge fact is a relational triple that consists of a head term, a tail term, and a relation between them, e.g., (patient, \textit{Desires}, health). 
We notice that some commonsense knowledge is represented by triples whose forms are consistent with the expression of language, e.g., (apple, \textit{Is}, red).\footnote{The concatenation of the triple ``\emph{apple is red}'' is close to the expression of language.}
We categorize such commonsense as plain commonsense knowledge. In contrast, there are some commonsense facts that are rarely expressed explicitly in natural language~\cite{gordon13} and their representation triples show inconsistency with language expression like (whale, \textit{AtLocation}, ocean). 
To describe this phenomenon, we define the extent of the inconsistency between a knowledge triple and language expression as the \emph{depth} of the triple. We call those triples with large depth as \emph{deep commonsense triples} and quantify this definition in this work. The knowledge facts implied by reasonable deep commonsense triples are termed as \emph{deep commonsense knowledge}.
Some typical examples are shown in Figure~\ref{fig:examples} to give an intuitive understanding.

Different knowledge triples are usually directly fed into models for processing in commonsense mining tasks. Prevailing methods adopt pre-trained language models for commonsense mining due to their impressive performance. The triples are fed into and then adapted to language models during the fine-tuning stage~\cite{KG-BERT, graph_ckbc}. 
Since language models learn the expression of language, for the plain commonsense, their triples can be easily adapted to language models owing to their consistency with language expression.
However, the deep commonsense facts are inherently inconsistent with language expression. As shown in Figure~\ref{fig:examples}, the knowledge fact (whale, \textit{AtLocation}, ocean) tends to be expressed in a more implicit way rather than directly saying ``\emph{whale is located at ocean}’’. We presume that it is hard to adapt such triples to appropriate language expressions and thus this inconsistency raises a challenge for pre-trained language models to handle them. 

However, little research has been done regarding this inconsistency between the knowledge representation and language models. In the prevailing methods, the triples are usually directly fed into language models ignoring the inconsistency between the input triples and the language expression. To better deal with the commonsense knowledge represented by the triples, it is essential to answer whether relying on adapting triples to language models during the fine-tuning stage can deal with all kinds of commonsense knowledge, especially the deep commonsense knowledge whose triple forms are inconsistent with language expressions.
%
Moreover, our experiments show that deep commonsense knowledge occupies a notable part of commonsense knowledge while prevailing methods have difficulty in handling them. In this paper, 
we make the first study about deep commonsense knowledge by probing the following problems.

\textbf{How to identify the deep commonsense triples?} The core of this question is to quantitatively define the depth of knowledge triples. A straightforward way is to manually grade the depth of triples. However, for one thing, manual annotation is usually expensive, for another, the evaluation standard varies for different individuals and thus hinders generalization. Therefore, developing automatic measurements for the depth of knowledge triple is indispensable. In our work, we develop two metrics to explicitly measure the depth and give a criterion to identify deep commonsense triples. 

\textbf{How to mine deep commonsense knowledge?} 
Deep commonsense knowledge is represented by the reasonable deep commonsense triples. However, our probing experiments show that conventional methods have difficulty in handling those triples with large depth, hindering the development of mining deep commonsense knowledge represented by them. 
To remedy this, we propose a novel method that directly processes the sentences that are indeed language expression in text form to mine the commonsense knowledge. The proposal encourages the model to discover the commonsense knowledge behind these sentences rather than relying on adapting triples to language expressions.

In summary, our contributions are as follows:
\begin{itemize}
\item We introduce the knowledge-language inconsistency problem and make the first endeavor to study the deep commonsense knowledge.

\item We propose a novel method with sentences as input to mine the knowledge distributed in sentences.
\item The proposed method significantly improves performance in mining the deep commonsense knowledge compared with baselines.

\end{itemize}

\section{Probing into Deep Commonsense Knowledge}\label{sec_probing}

Although we introduce the concepts of the depth of knowledge triples and deep commonsense knowledge, how to explicitly measure the depth and quantitatively define the deep commonsense triples remain unexplored. In this section, we investigate these essential problems and study the effect of deep commonsense knowledge through \textbf{C}ommonsense \textbf{K}nowledge \textbf{B}ase \textbf{C}ompletion~(CKBC) task since it is a classic task for commonsense knowledge mining.
\subsection{Preliminaries}
CKBC aims to distinguish the reasonable knowledge triples from the unreasonable ones. Following previous work \cite{li2016commonsense}, we treat this task as a binary classification problem to label high-quality triples as positive while labeling the unreasonable ones as negative. In our probing experiments, we also adopt the evaluation dataset introduced by~\citet{li2016commonsense}, where the reasonable triples are from the crowd-sourced Open Mind Common Sense entries and the unreasonable triples are generated by negative sampling. 
We denote this evaluation dataset as OMCS for short.
In light of the impressive performance of pre-trained language models, we employ pre-trained language models to tackle the CKBC task. A natural way is to directly feed the concatenation of triples into language models and then additionally learn a new classification layer in the fine-tuning stage. In this paper, we regard such a treatment as the conventional method due to its simplicity and effectiveness, and study its performance on the CKBC task.

\subsection{Measurement of Depth}

\begin{figure}[t]
\centering
\subfigure[Depth rank and perplexity given by GPT-2.]{
\begin{minipage}[t]{1.0\linewidth}
\centering
\includegraphics[width=7.5cm]{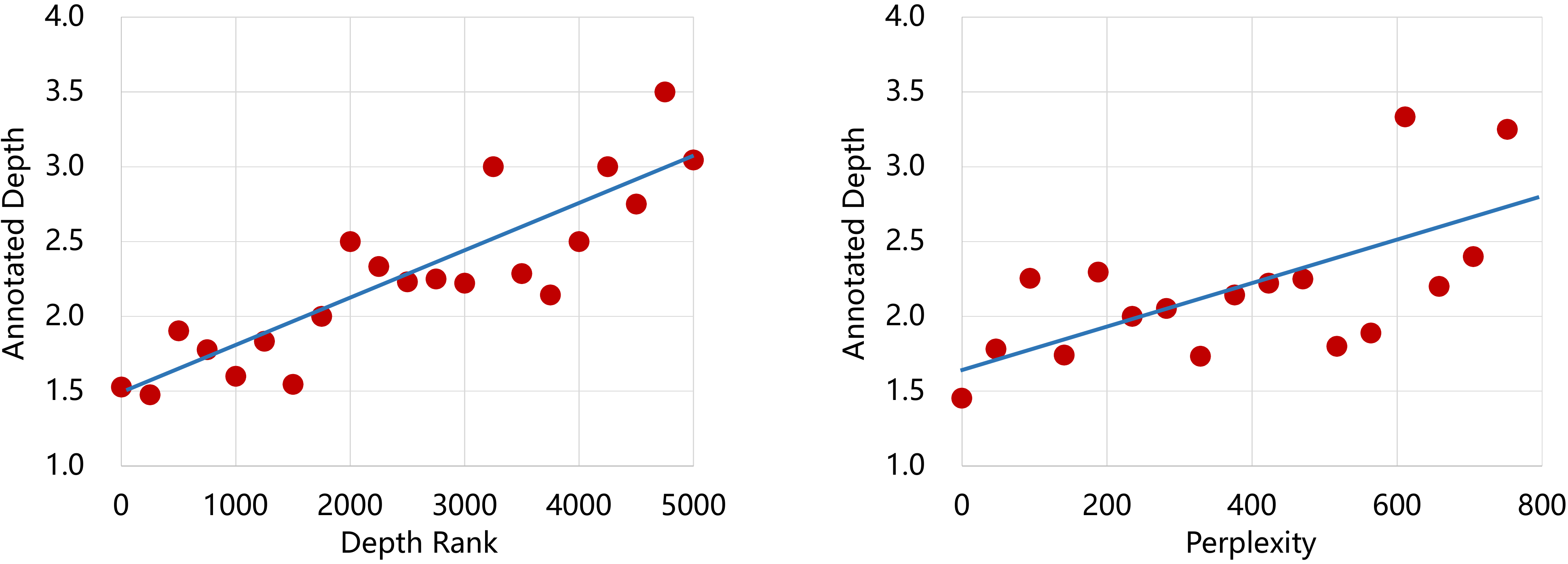}
\end{minipage}%
\label{fig_rank_gpt}
}%
\\
\subfigure[Depth rank and perplexity given by Transformer-XL.]{
\begin{minipage}[t]{1.0\linewidth}
\centering
\includegraphics[width=7.5cm]{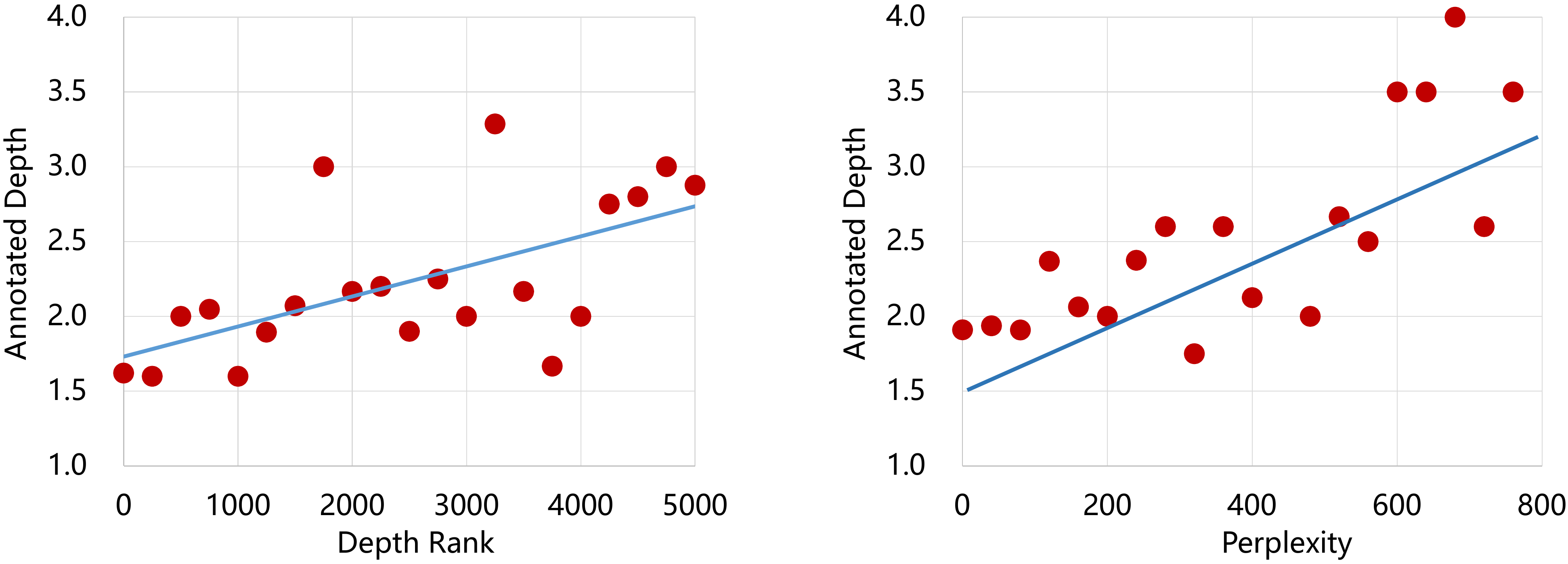}
\end{minipage}%
\label{fig_ppl_gpt}
}%
\centering
\caption{The correlation between human annotated knowledge depth and automatic measurements. The blue line is the trend line of human annotated results.}
\label{fig_correl}
\end{figure}

Since the depth reflects the extent of inconsistency between a triple and the expression of language, and language models are designed to learn the expression of language, we attempt to use metrics derived from the language model to measure the depth. 
Before calculating these metrics, triples are first converted to natural language with simple templates.
We do not excessively design these templates and are tolerant of some grammatical issues of the converted sentences since simple templates preserve the original form of a knowledge triple and thus directly indicate the depth of the triples. The two metrics are:

\paragraph{Depth rank:} 
    Given a sentence $S = \{h_{1:m}, r_{1:n}, t_{1:k}\}$ converted from a triple, where $m$, $n$ and $k$ denote the number of tokens in head term, relation and tail term, respectively, all the tokens in this sentence are auto-regressively fed into language models and each token will obtain a prediction rank. We define the depth rank as:
    \begin{equation}
    \small 
        {\rm DepthRank}(S) = \frac{ \sum_{i=1}^k {\rm Index} (t_i| h_{1:m}, r_{1:n}, t_{<i} )}{k}
    \end{equation}
    where ${\rm Index}(\cdot|\cdot) $ is the rank index of the correct tail token in the model prediction result given the head and relation tokens. 
    The depth rank reflects the preference of language models for generating words.
\paragraph{Perplexity:} The second metric is the perplexity of the sentence converted from a triple, obtained from pre-trained language models.

We assume that the larger the depth rank and the perplexity is, the deeper the knowledge triple is. 
To verify that the depth of knowledge fact is proportional to the introduced metrics, we manually annotate 600 reasonable knowledge triples from Open Mind Common Sense regarding their depth. 
We adopt a 4-point scale where 1 represents the shallowest knowledge triple and 4 represents the deepest one to encourage the annotators to take a stand on whether the triple is declined to shallow or deep. More annotation details can be found in Appendix~A. We divide the triples into multiple groups according to their depth rank and perplexity, and calculate the human-annotated depth of each group. To show the generality, we adopt two pre-trained language models, i.e., GPT-2~\cite{radford2019better} and Transformer-XL~\cite{dai2019transformer}.
The correlations of human-annotated results and automatic criteria are shown in Figure~\ref{fig_correl}. We observe that both metrics demonstrate a strong correlation with human-annotated results. 
It indicates that the depth rank and perplexity can be utilized as proxies for measuring the depth of a knowledge triple. Particularly, the depth rank calculated by GPT-2 shows the highest correlation coefficient of 0.67 with human annotations. For convenience, we use this metric as a standard criterion to assist the following study.
\begin{figure}[t]
\begin{center}
\includegraphics[width=0.9\linewidth]{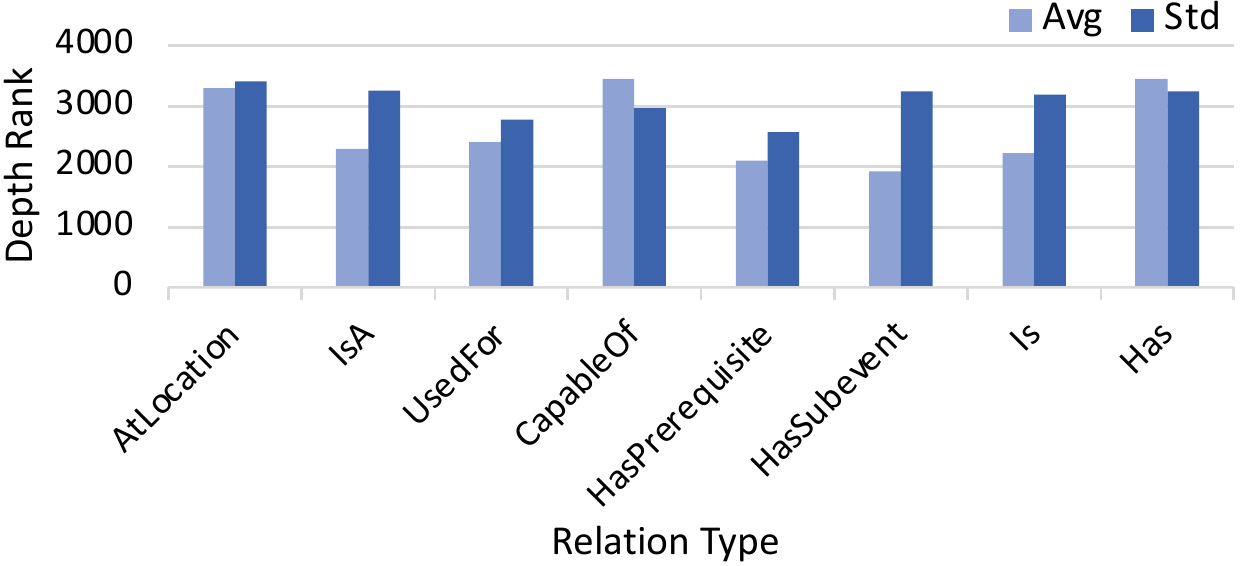}
\end{center}
\caption{Depth rank average and standard deviation of different relation types in the OMCS dataset. While relations close to language expressions tend to have a low average depth rank, the variance is high since the depth is also affected by head and tail terms.}
\label{fig_omcs_rank}
\end{figure}
\vspace{-0.1cm}

\subsection{Effect of Relation Type on Depth}
We also explore the correlation between depth rank and relation type for a comprehensive understanding of the deep commonsense knowledge. 
In more detail, we group triples in OMCS dataset by relation type and compute the average and standard deviation of depth rank for each group in Figure~\ref{fig_omcs_rank}. The average depth rank varies among different relations, which suggests that the depth of a triple is affected by the relation type to some extent, e.g., relations close to language expression like \textit{Is} tend to have a low average depth rank. 
However, the observed high variance in the depth reveals that those commonsense triples with the same relation can significantly differ in depth. Such difference demonstrates that head and tail terms also have a big impact on the depth of a triple, e.g., the triple (whale, \textit{AtLocation}, ocean) is deeper than (Eiffel Tower, \textit{AtLocation}, Paris).


\begin{figure}[t]
\begin{center}
\includegraphics[width=0.88\linewidth]{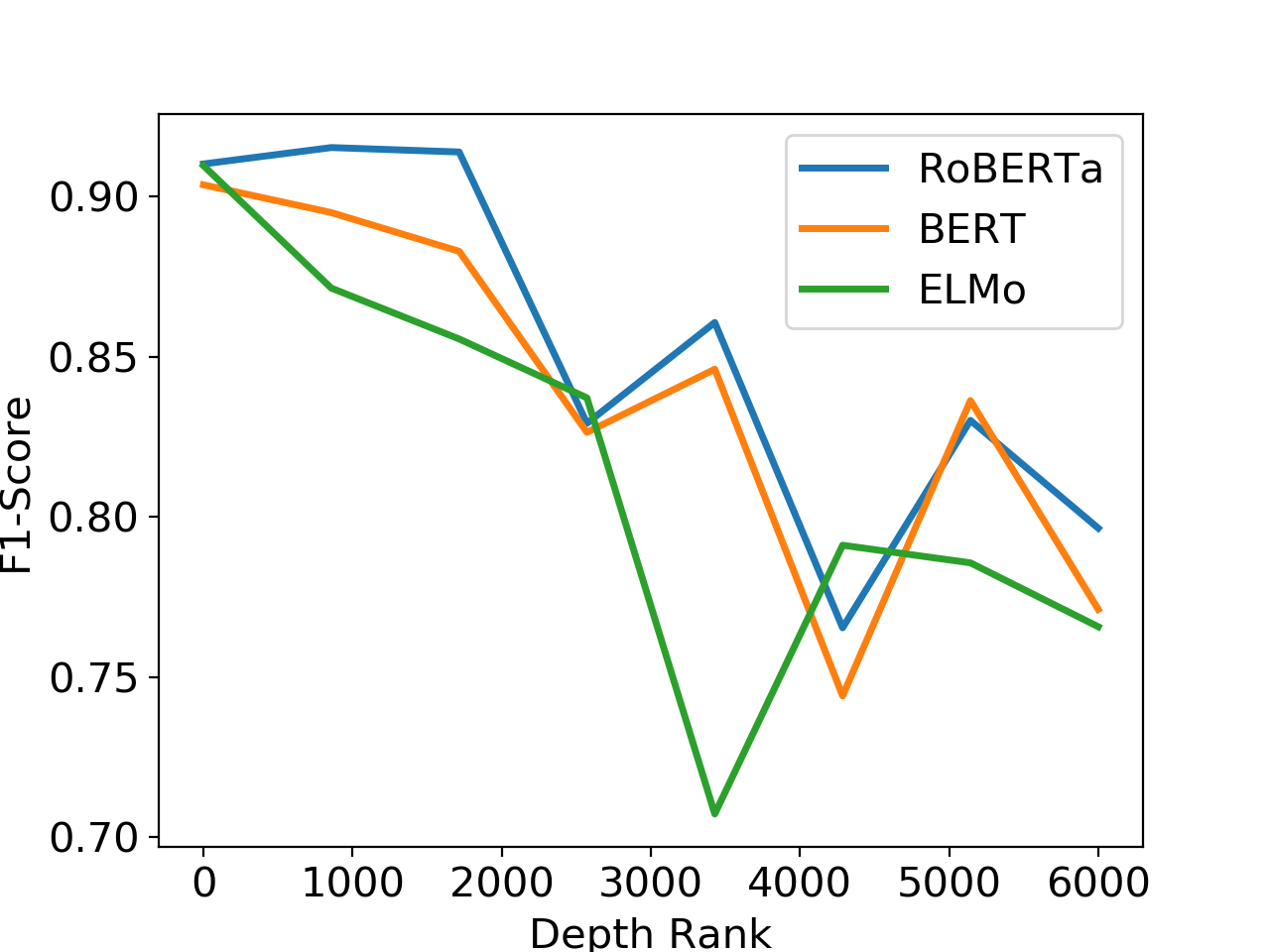}
\end{center}
\caption{Performance of typical pre-trained language models on the entire OMCS test set. We observe a significant performance drop around the rank of 2,000.}
\label{fig_whole_omcs_test}
\end{figure}
\subsection{Significance of Deep Commonsense}\label{deep_triple_def}
We then take a step further to investigate the effect of the depth through CKBC task regarding the performance of using pre-trained language models. We adopt three pre-trained language models widely used in classification tasks, i.e., ELMo~\cite{peters2018deep}, BERT~\cite{devlin2019bert} and RoBERTa~\cite{liu2019roberta}. 
Detailed setup is included in Appendix~B.
The performance curve on different groups sorted by depth rank is shown in Figure~\ref{fig_whole_omcs_test}. We observe that the performance varies on different depth of knowledge facts. In general, the performance deteriorates as the depth increases.

As deep commonsense knowledge can be represented by the deep knowledge triples, the above phenomenon discloses that the prevailing methods using pre-trained language models have difficulty in dealing with deep commonsense knowledge. Such methods take the concatenated triple as the input of language model. During the fine-tuning stage, the model actually learns to narrow the gap between the triple and the corresponding language expression. For the plain commonsense, the language model can easily adapt them to language form and handle them with facility. However, for those triples with high depth rank, it is relatively difficult to fit them to appropriate language form. These knowledge triples remain unfamiliar and intractable form for language models to handle. As a result, the model is confused in tackling the deep commonsense triples, and accordingly the performance declines significantly as the depth increases.


Particularly,  we observe that the performance significantly drops on the rank close to $2,000$. Therefore we take the depth rank $2,000$ given by GPT-2 as a boundary to quantitatively identify the deep commonsense triples. That is, the knowledge triples with depth rank larger than $2,000$ are categorized as deep commonsense triples in our work.

\begin{figure}[t]
\centering
\subfigure[Proportions of deep commonsense knowledge in the OMCS dataset.]{
\begin{minipage}[t]{0.45\linewidth}
\centering
\includegraphics[width=3.0cm]{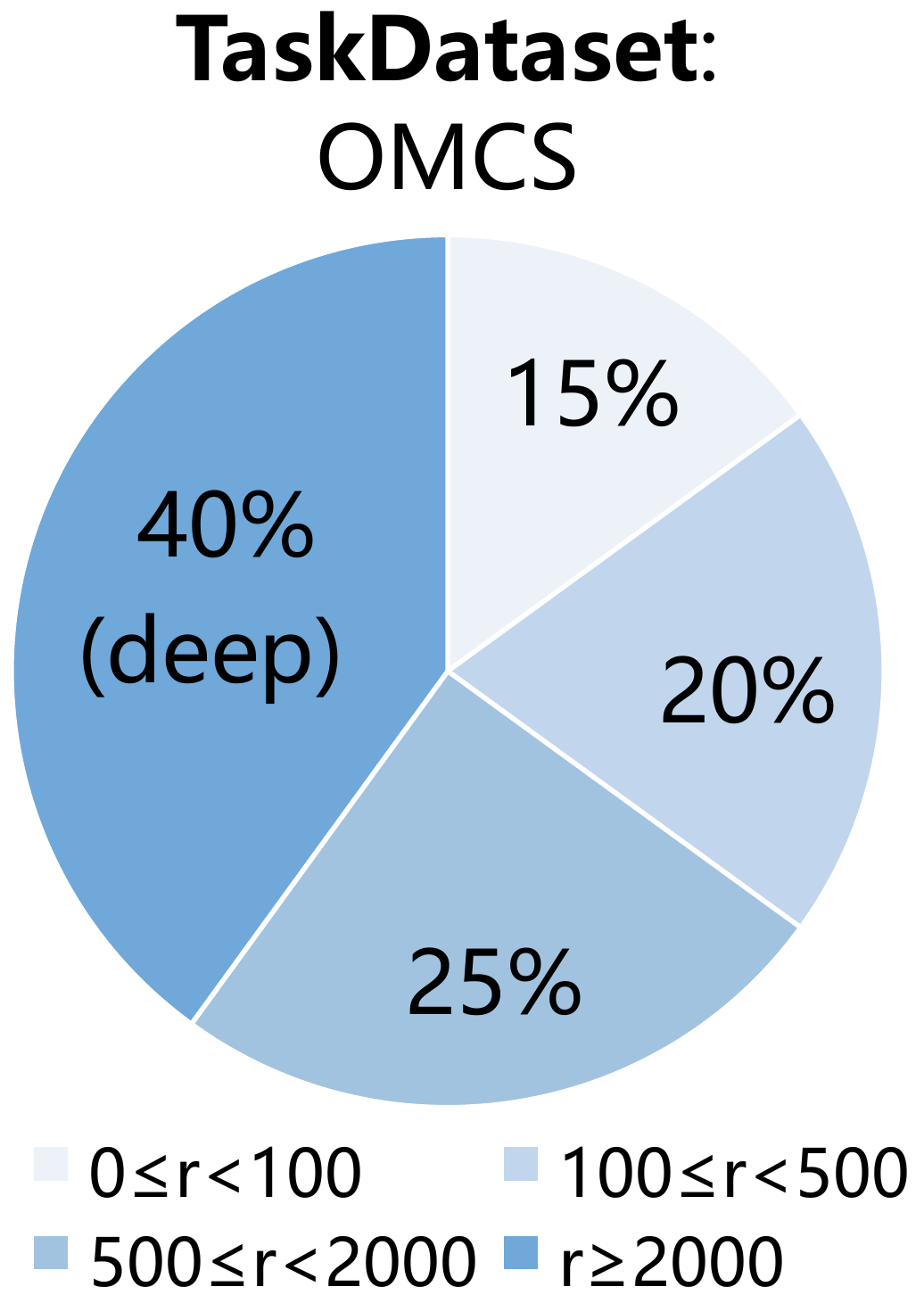}
 \label{fig_prop1}
\end{minipage}%
}%
\quad
\subfigure[Proportion of deep commonsense in the knowledge base ConceptNet.]{
\begin{minipage}[t]{0.45\linewidth}
\centering
\includegraphics[width=3.0cm]{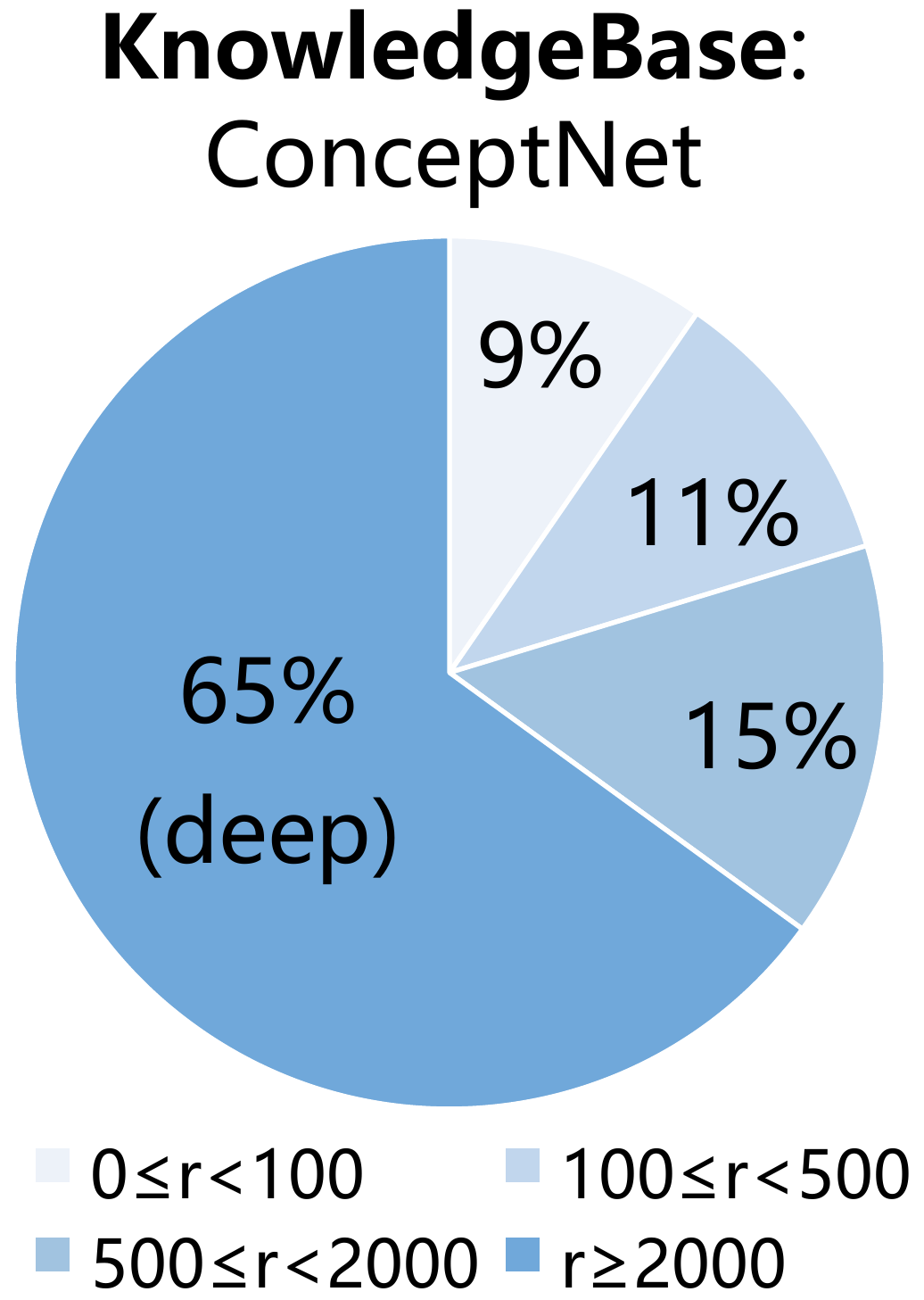}
\end{minipage}%
\label{fig_prop2}
}%
\centering
\caption{The distribution of different depth rank in the knowledge base ConceptNet and the OMCS dataset for CKBC task. r denotes the depth rank.}
\label{fig_prop}
\end{figure}

Along with the challenge posed by deep commonsense knowledge, another question arises: does deep commonsense knowledge really occupy a notable part of the commonsense knowledge? We then investigate the depth of triples in the CKBC task dataset and the commonsense knowledge base ConceptNet~\cite{ConceptNet}. The statistical result for task dataset reveals the occurrence frequency of deep commonsense knowledge in downstream tasks, and the result for knowledge base reflects the proportion of deep commonsense knowledge in real commonsense knowledge.  
We first sample $2,000$ entries in OMCS evaluation dataset and compute their depth ranks. The proportions of different depth ranges are shown in Figure~\ref{fig_prop1}. Deep commonsense knowledge occupies about 40\% of the samples, which is a significant part. We further randomly sample 10k knowledge triples in the ConceptNet and measure their depth. According to Figure~\ref{fig_prop2}, the plain commonsense occupies a small part of knowledge while about 65\% knowledge is actually deep commonsense knowledge. These results suggest that deep commonsense plays a role that can not be overlooked in the research of commonsense knowledge.

\section{Mining the Commonsense Knowledge Distributed in Sentences}

The analysis in Section~\ref{sec_probing} demonstrates that conventional methods heavily rely on adapting triples to language expression, leading to sub-optimal performance when dealing with deep commonsense knowledge. Therefore we attempt to provide sentences that are exactly language expression as the input of language models, which helps elude the inconsistency problem exposed in conventional methods. We speculate that deep commonsense can be implicitly inferred from sentences rather than learning from the artificial triples. For instance, given a reasonable triple (take a bath, \textit{HasSubevent}, wash hair), the sentences "\emph{He went to bath to \textbf{take a bath} only to find the shampoo was used up}" and "\emph{Finally rinse hair completely, then \textbf{wash hair} with some shampoo}" provide potential evidence for correctly identifying the above triple. %
Therefore we propose a novel method which identifies reasonable triples through mining the knowledge distributed in sentences related to triples. 

\begin{figure}
    \centering
    \includegraphics[width=0.99\linewidth]{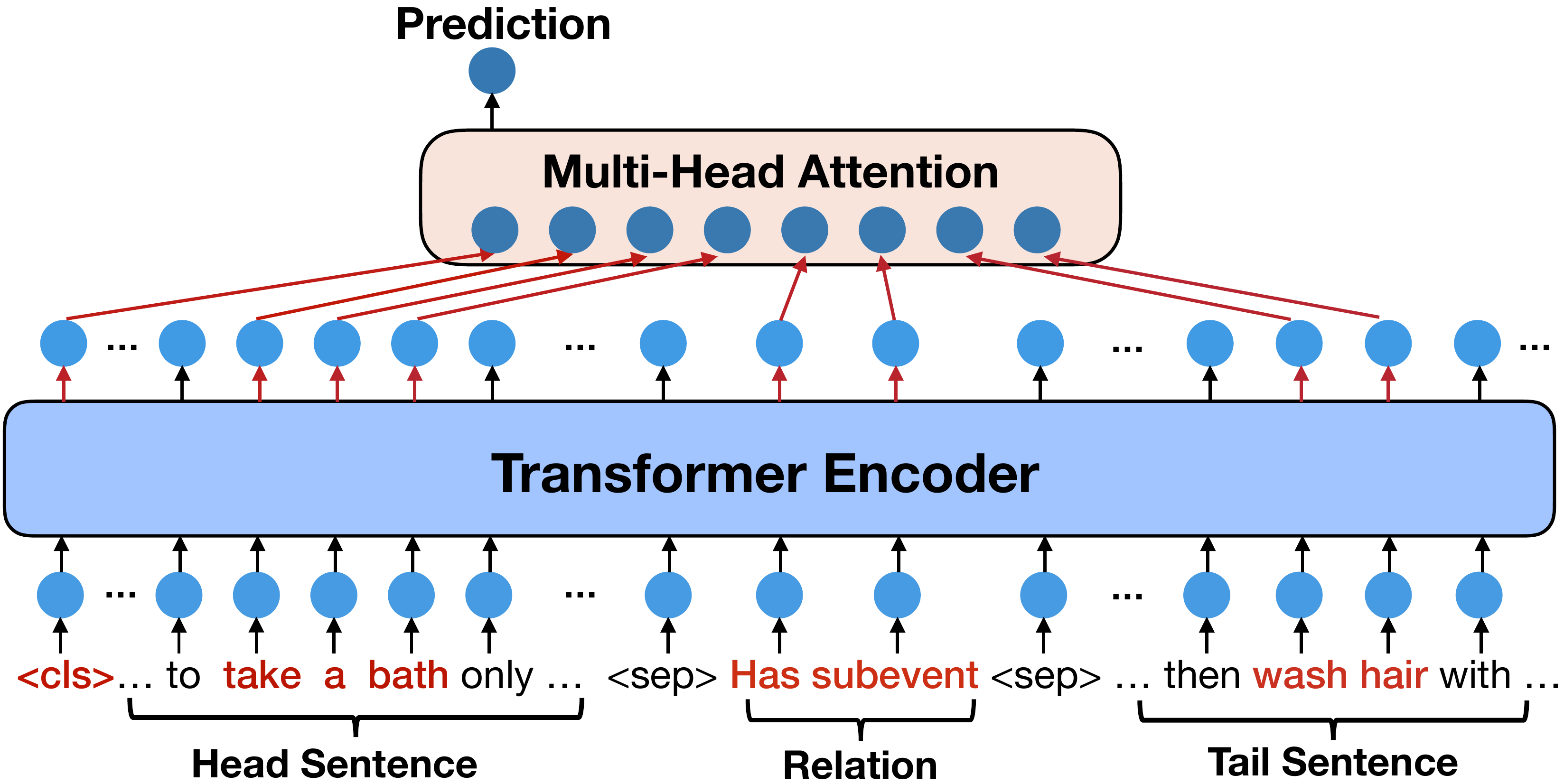}
    \caption{The overview of the our model architecture. The head sentence and tail sentence denote the sentences containing head term and tail term, respectively.}
    \label{fig:train_overview}
\end{figure}

Concretely, we select two sentences that contain head term and tail term respectively and also have the most word overlaps. They serve as the contexts of the corresponding triple to provide potential clues and enrich the semantic meanings of head and tail terms. A special case is that the head and tail terms appear in one sentence. In such a situation, the two input sentences are identical. We discuss the more general case in this work.
Instead of forcefully adapting the deep knowledge triple to the corresponding language expression, which has been demonstrated ineffective in our previous analysis, our proposal makes classification on the sentences that are exactly language expressions and are readily processed by language models. Thus, the model can discover the implicit deep knowledge behind these sentences with facility. An overview of our model is shown in Figure~\ref{fig:train_overview}.

Formally, given a triple $x = (h,r,t)$, we denote the sentence containing the head $h$ as $s^h$ and the sentence containing the tail $t$ as $s^t$. Intuitively, the sentence pairs with the most relevant contexts will provide more beneficial information about the relation between head and tail. Therefore, we select the sentence pair $(s^h, s^t)$ with most word overlap after removing stop words from raw texts. Here we use the raw texts from BOOKCORPUS~\cite{zhu2015aligning}. Meanwhile, we convert the relation to a phrase, e.g., ``CapableOf'' is rephrased as ``capable of''.  Each sentence pair and the rephrased relation $R$ form a new triple $(s^h, R, s^t)$. We additionally add a \texttt{<sep>} token between the relation phrase and sentences to distinguish their boundaries and a \texttt{<cls>} token is appended to the beginning of the sequence. The resulting sequence is denoted as $X$ and serves as the input of our model.

We feed the input sequence $X$ into a pre-trained language model which in our case is RoBERTa~\cite{liu2019roberta} for encoding. We extract the token representations of $\texttt{<cls>}$, $h$, $R$, and $t$ in the last layer and denote them as $E^{\texttt{<cls>}}$, $E^h$, $E^r$ and $E^t$, respectively. With the interactions provided by multi-layer encoder, we assume that these representations incorporate rich semantic information from two context sentences with respect to the original triple $(h, r, t)$. We concatenate them as $\hat{E}$ and perform the standard multi-head self-attention operation~\cite{vaswani2017attention} on $\hat{E}$, resulting in the final hidden state $H$ of input $\hat{E}$. The hidden state corresponding to $\texttt{<cls>}$ is denoted as $H^{\texttt{<cls>}}$ and is followed by a feed forward layer to produce an output distribution over target labels. This procedure can be formulated as:
\newenvironment{smallalign}{\par\small\align}{\endalign}
\begin{smallalign}
&E_0={\rm Emb\_Layer}(X) \\
&E_l={\rm Transformer\_Layer}(E_{l-1}), \forall l\in[1, n]\\
&\hat{E}=E_n^{\texttt{<cls>}} \oplus E_n^{h} \oplus E_n^r \oplus E_n^{t} \\
&H={\rm Self\_Attn}(\hat{E}) \\
&P={\rm Softmax}(WH^{\texttt{<cls>}})
\end{smallalign}
where $n$ is the number of encoder layers, $W \in \mathbb{R}^{2 \times d_v}$ is a matrix parameter and
$P \in \mathbb{R}^{2}$ is the binary probability distribution.

Since there are different available sentences containing the head $h$ or the tail $t$, we select multiple sentence pairs 
to make a comprehensive prediction. 
Specifically, we retrieve a sentence set $\mathcal{S}_H = \{{s_i^h}\}_{i=1}^{|S_H|}$ for head $h$ and a sentence set $\mathcal{S}_T = \{{s_j^t}\}_{j=1}^{|S_T|}$ for tail $t$. 
We then select top $K$ sentence pairs $S_{pair} = \{(s_k^h,s_k^t) \}_{k=1}^{K}$ with most word overlap, establishing an evidence set $\mathcal{D}$ with size $K$ for assessing the triple $x$: 
\begin{equation}
\small
    \mathcal{D} = \{(s_k^h, R, s_k^t) | (s_k^h, s_k^t) \in S_{pair}\}_{k=1}^K
\end{equation}

For each triple $x = (h, r, t)$, the evidence set $\mathcal{D}$ contains $K$ pairs $(s^h, R, s^t)$, resulting in $K$ prediction results which is denoted as $\{{p^k}\}_{k=1}^{K}$ in the following description. 
The model is optimized via minimizing the average of the negative log-likelihood over the $K$ sequences:  
\begin{equation}
\small
    \mathcal{L} = - \frac{1}{K}\sum_{k=1}^{K} Y\log (p^k)
\end{equation}
where $Y$ is he gold label of the original triple. 

During the inference stage, we ensemble these $K$ prediction results to make a final prediction. We denote $p_0^k$ and $p_1^k$ as the $k$-th probability being fictitious and true, respectively. We develop three strategies to assemble temporary prediction results:

\begin{itemize}
\item  \textbf {Avg-Prediction} The average probability reflects the comprehensive reasonability of the triple.
\vspace{-0.1cm}
\begin{equation}
\small
\hat Y = {\rm{argmax}}{ (\frac{1}{K} \sum_{k=1}^{K} p_{0}^k , \frac{1}{K} \sum_{k=1}^{K} p_{1}^k)}
\end{equation}
\item \textbf{Max-Prediction} The max probability represents the most confident prediction of the triple.
\begin{equation}
\small
\hat Y = {\rm{argmax}}( {\rm max}(  \{p_{0}^{k}\}_{k=1}^{K} ), {\rm max}(\{p_{1}^{k}\}_{k=1}^{K}  ))
\end{equation}
\item \textbf{Vote-Prediction} The vote result reflects the validity of a triple in most cases.
\vspace{-0.1cm}
\begin{small}
\begin{align}
N_0 &= \sum_{k=1}^K ( \mathbf{1}( {\rm argmax}(p_0^k, p_1^k ) = 0 ) \\
N_1 &= \sum_{k=1}^K ( \mathbf{1}( {\rm argmax}(p_0^k, p_1^k ) = 1 ) \\
\hat Y &= {\rm argmax} (N_0, N_1) 
\end{align}
\end{small}
where $\mathbf{1}(\delta)$ is an indicator function giving result $1$ if $\delta$ is true, otherwise $0$.
\end{itemize}

\section{Experiments}
\subsection{Datasets}

We use a total of 500k triples from ConceptNet as positive examples of training data. Among them, 100k are from the Open Mind Common Sense entries.
The total 500k positive triples along with the 500k negative triples generated by negative sampling serve as our training set. The first evaluation dataset is established by \citet{li2016commonsense} from Open Mind Common Sense and is denoted as \textbf{OMCS} in this work. It has 2,400 examples in the development set and the test set respectively. Since the OMCS evaluation dataset has no distinction of the depth of triples, and to enlighten the research of deep commonsense knowledge, we construct a new evaluation dataset that consists of only deep commonsense triples and human-annotated labels. It has 2,000 examples in the development set and the test set respectively. The construction details are included in Appendix~C. 
We denote this dataset that contains \textbf{D}eep \textbf{C}ommon\textbf{S}ense \textbf{K}nowledge as \textbf{DCSK} for short. We conduct experiments on these two datasets.

\subsection{Experimental Settings}

We implement our proposed model based on the architecture of RoBERTa. In detail, we use the transformer architecture with 24 layers. It contains 16 self-attention heads and the hidden dimension is 1,024. For our self-attention block stacked on the transformer blocks, we use 8 self-attention heads. The hidden dimension is bounded with that of transformer blocks thus it is 1,024. We use the pre-trained parameters of RoBERTa$_{large}$ to initialize this part of our model. The parameters of the upper part of our architecture, namely the self-attention block and the classifier are randomly initialized. The learning rate is set to $1e^{-5}$. We train the model for a total of $24$k steps with Adam optimizer~\cite{Adam}. Additionally, we set the number of context pairs $K$ as $3$ and employ the inference strategy of Avg-Prediction since we empirically find it works best on the development set. Detailed discussion about the hyper-parameter $K$ and different inference strategies are included in Appendix~D and Appendix~E, respectively.

\subsection{Baselines}
We implement several commonly-used baseline methods for CKBC:
\begin{itemize}
    \item DNN-Avg, DNN-LSTM: DNN-Avg~\cite{li2016commonsense} averages the token embeddings of the triple for representation; DNN-LSTM~\cite{li2016commonsense} utilizes a bi-directional LSTM for representing tokens. The representations are followed by a multi-layer perceptron~(MLP) classifier for both models.
    \item ELMo, RoBERTa, BERT: ELMo~\cite{peters2018deep} takes the representation of the sentence converted from triples to make classification. RoBERTa and BERT~\cite{devlin2019bert} are fine-tuned on our task with triples as input.
\end{itemize}

\subsection{Overall Performance} 
To verify the effectiveness of mining deep commonsense knowledge, we select the deep commonsense triples of OMCS test set according to the definition of deep commonsense triples~(see Section~\ref{deep_triple_def}) to form a sub-test set. The performance of models on this test set is shown in Table~\ref{tab:omcs_test}. We notice that DNN-based methods perform well on this dataset. We conjecture that some triples in the test set are rewordings of triples in the training set, which is consistent with the observation in previous work~\cite{jastrzebski2018commonsense}, thus simple methods based on word embedding similarity can also achieve good performance. Although pre-trained language models are considered to contain commonsense knowledge to some extent~\cite{LAMA}, the conventional methods with pre-trained language models heavily rely on narrowing the gap between triple form to language expression during the fine-tuning stage and show weakness in handling deep commonsense knowledge.
In contrast, the proposed method significantly outperforms the baselines, which validates its advantage over conventional methods.

\begin{table}
\begin{center}
\begin{tabular}{c|c|c|c}
\toprule 
       OMCS (deep) & Precision &Recall & F1-score   \\
\midrule
    DNN-Avg &  77.13& 94.16 &84.80 \\
    DNN-LSTM  & 77.01 & 93.51  &  84.46\\
    \midrule
    ELMo & 77.78 &88.64 & 82.85 \\
    BERT &75.73 &92.21 &83.16 \\
    RoBERTa &75.52 &93.18 &83.43 \\
    Our Model &\textbf{80.06} &\textbf{96.42} & \textbf{87.48} \\
\bottomrule
\end{tabular}
\caption{Performance on the deep commonsense triples of OMCS test set. Best results are shown in bold.}
\label{tab:omcs_test}
\end{center}
\end{table}

The performance on the DCSK dataset is shown in Table~\ref{tab:cmk_test}. Compared to Table~\ref{tab:omcs_test}, we observe a significant performance drop on all the methods. The main reason is that the depth of the triples in the DCSK dataset is substantially larger than that in the OMCS dataset, which means the DCSK dataset is more challenging.
We observe that simple neural networks perform poorly with very low F1-score. These methods largely depend on embedding similarity and thus cannot recognize the valid deep commonsense triples which are novel items, leading to a low recall and further yielding poor performance. 
The approaches using pre-trained models including our model achieve significant improvement compared with DNN-based models, indicating that it is advantageous to employ pre-trained models that obsess commonsense knowledge background.

\begin{figure}[t]
\begin{center}
\includegraphics[width=0.9\linewidth]{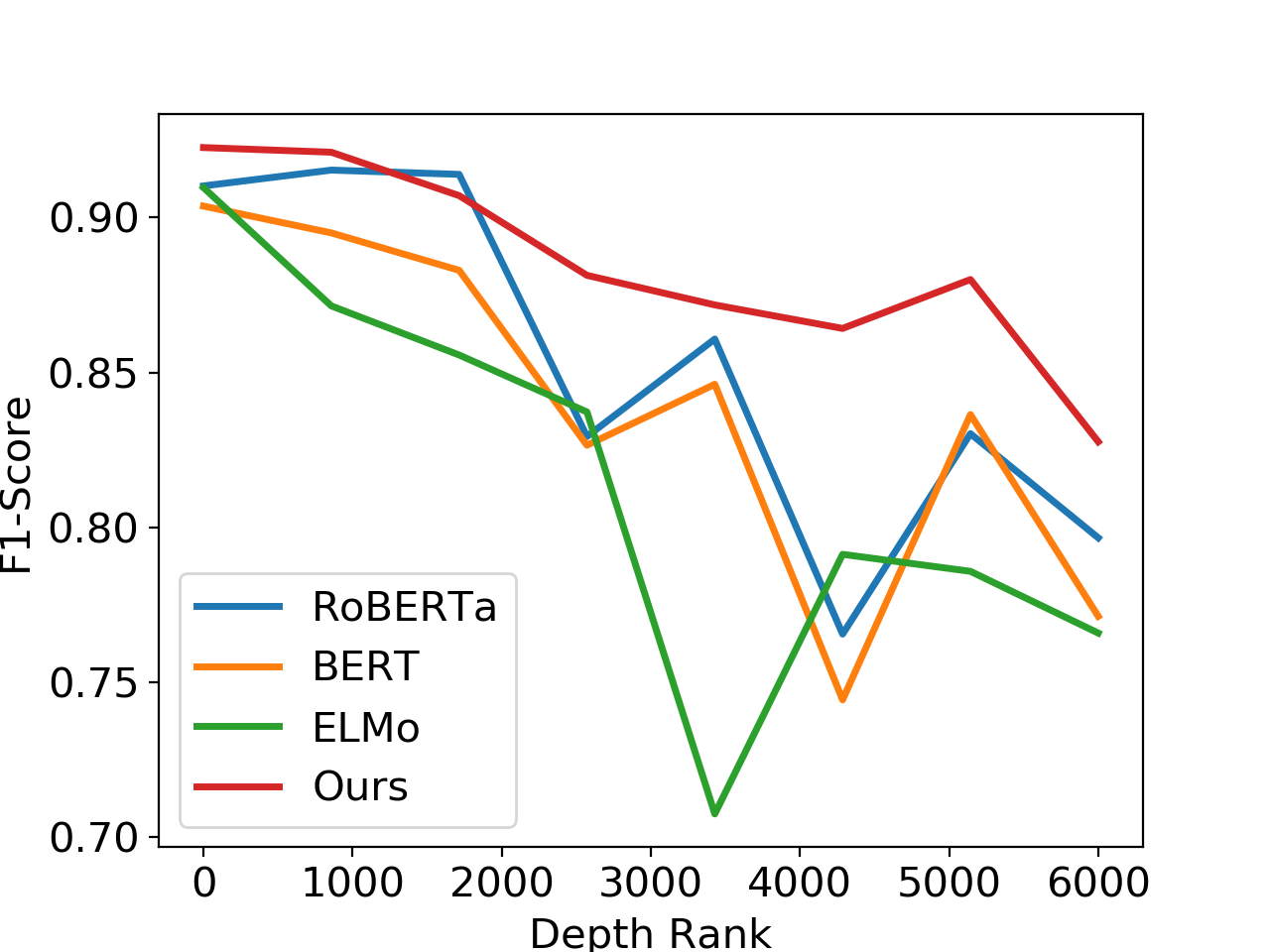}
\end{center}
\caption{Performance of conventional methods using pre-trained language models and our method on the entire OMCS test set.}
\label{fig_deep_omcs_test}
\end{figure}

Our proposal achieves the best performance in these two datasets and we speculate the improvement is two-fold. First, we alter the conventional input to sentences whose forms are consistent with language expression. Thus it is more beneficial and effective for language models to process the input, alleviating the reliance of conventional methods on the representation form of knowledge. Second, the sentences also serve as the contexts for both head term and tail term. The contexts enrich their semantic meanings and provide potentially valuable clues to capture the deep commonsense knowledge.
 \begin{table}
\begin{center}
\begin{tabular}{c|c|c|c}
\toprule 
       DCSK & Precision  & Recall & F1-score \\
\midrule
    DNN-Avg &  \textbf{53.75}& 11.88 & 19.46\\ 
    DNN-LSTM & 39.79  & 20.72 & 27.25\\
    \midrule
    ELMo &48.10 &43.65 &45.76 \\
    BERT & 47.59& 62.85& 54.17\\
    RoBERTa &46.67 &65.05 &54.35 \\
    Our Model &51.40 &\textbf{68.09} &\textbf{58.58} \\
\bottomrule
\end{tabular}
\caption{Performance on our proposed DCSK test set consisting of only deep commonsense triples. Best results are shown in bold.}
\label{tab:cmk_test}
\end{center}
\end{table}
In addition to the experiments on pure deep commonsense triples, we also perform experiments on the entire OMCS test set. 
The test set is divided into multiple groups according to their depth ranks to see the performance of different groups. The results of conventional methods and our method are shown in Figure~\ref{fig_deep_omcs_test}. The proposal shows a clear advantage over baselines when dealing with deep commonsense knowledge and also slightly outperforms RoBERTa when the depth rank is relatively low. It reveals that the proposal will not hurt performance when handling plain commonsense knowledge.

\subsection{Case Study}

\begin{figure}[t]
    \centering
    \includegraphics[width=0.95\linewidth]{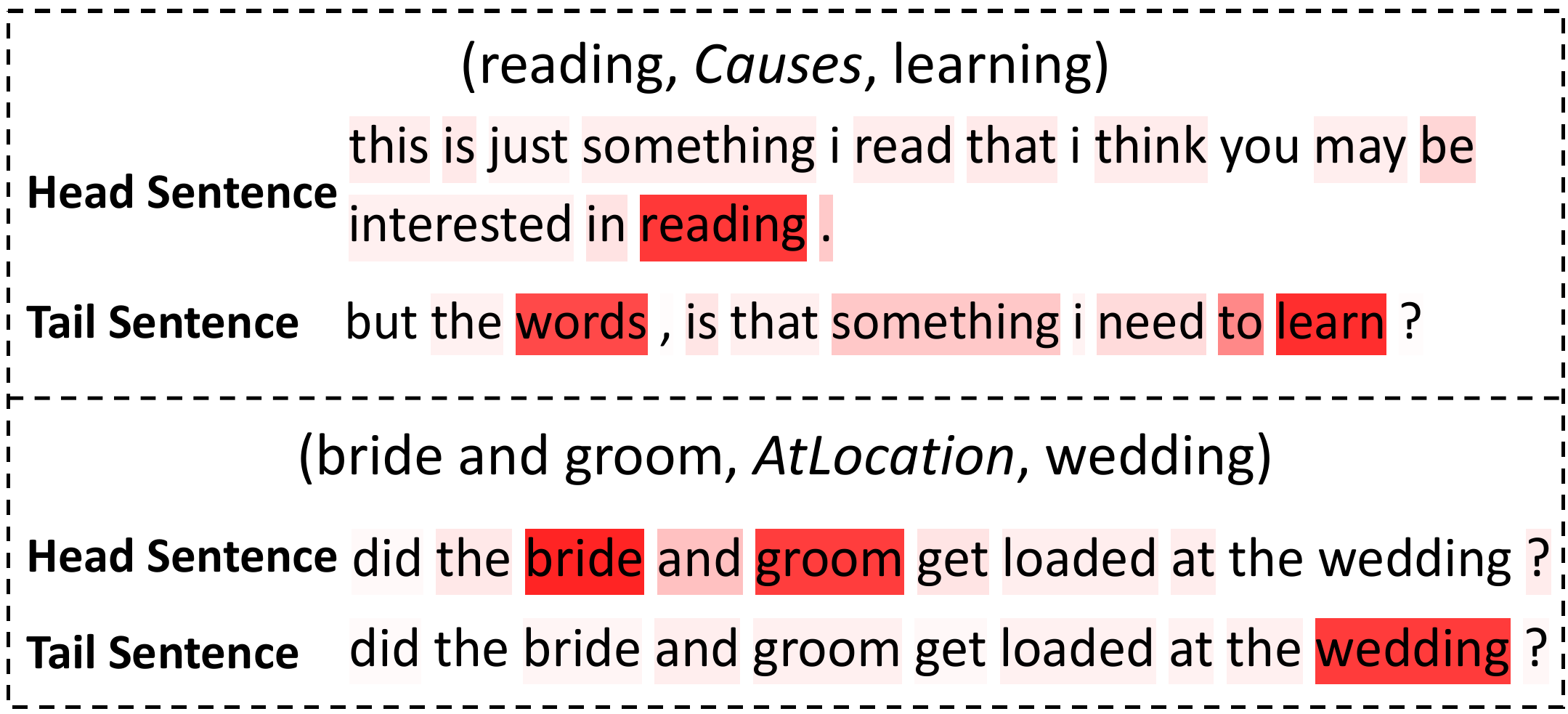}
    \caption{Attention heatmap visualizations of the proposal mining commonsense knowledge distributed in sentences. Note that RoBERTa fails to predict correctly in both cases, while the proposal succeeds.}
    \label{fig:case_2}
\end{figure}

In Figure~\ref{fig:case_2} we show some reasonable commonsense knowledge triples that RoBERTa fails to recognize while the proposed method correctly labels them as positive. Since the final representation of the token \texttt{<cls>} is used for classification, we trace its attention weights back to the tokens of the two input sentences and highlight the words with large attention weights. Regarding the first case, "words" can not be directly learned, and actually it is the process of reading the words that helps people learn something. We hypothesize that the token "words" in the second sentence provides a strong indication to build the connection to the head term "reading" in the first sentence, and we also find that it attracts more attention. These two sentences together hint at the validity of the corresponding knowledge triple. The second example shows the special case when the head term and tail term appear in one sentence, resulting in two identical input sentences. Our model can attend to different useful parts and then makes classification on two same contexts.

\section{Related Work}

Commonsense mining is an active research area. 
A typical task is knowledge base completion~\cite{socher2013reasoning,yang2014embedding,wang2015knowledge,DBLP:conf/naacl/NguyenNNP18}, which distinguishes the valid knowledge facts from the fictitious ones.
\citet{li2016commonsense} employ DNN-based methods for this task, but \citet{jastrzebski2018commonsense} prove such methods have difficulty in mining novel commonsense knowledge. \citet{davison2019commonsense} attempt to use unsupervised method and they convert the triples into sentences via hand-crafted templates and feed them into pre-trained language models to evaluate the validity.  
However, such a method heavily relies on the coherence of the sentences. 
Apart from the classification-based methods, recent researchers also explore generative models to mine commonsense knowledge. 
\citet{saito2018commonsense} and \citet{sap2019atomic} train encoder-decoder models to generate commonsense knowledge by using ConceptNet~\cite{ConceptNet} and ATOMIC~\cite{sap2019atomic} as underlying KBs, respectively. \citet{bosselut2019comet} take advantage of pre-trained language models to generate the tail term given a fixed head term and a specific relation, contributing to producing more new knowledge triplets. 
Nonetheless, all the previous methods employing pre-trained language models take the relational triples as input and thus heavily rely on the triple form of knowledge facts.

Our work also relates to the work exploring the knowledge implied in pre-trained language models~\cite{knowledge_in_contextual1, knowledge_in_contextual2}. \citet{LAMA} make a comprehensive study to seek to what extent pre-trained language models store world sense and commonsense knowledge. 
Particularly, \citet{commonsense_in_lm1} and \citet{commonsense_in_lm2} focus on evaluating the commonsense revealed by pre-trained language models, indicating that pre-trained language models include plenty of commonsense knowledge. 
However, they treat all the commonsense knowledge on an equal basis ignoring their inherent types. In this paper, we first make distinctions in plain commonsense and deep commonsense and further demonstrate that mining deep commonsense knowledge with pre-trained language models need more elaborate techniques.

\section{Conclusion}

In this paper, we introduce a phenomenon that the triple forms of some commonsense facts are inconsistent with language expressions and put forward the issue of deep commonsense knowledge. We conduct extensive experiments to understand deep commonsense knowledge and show that deep commonsense knowledge occupies a notable part of knowledge while conventional usage of pre-trained language models fails to capture it effectively. We then propose a novel method to directly mine commonsense knowledge from sentences, alleviating the inconsistency problem. Experiments demonstrate that our proposal significantly outperforms baseline methods in mining deep commonsense knowledge.

\bibliography{anthology,custom}
\bibliographystyle{acl_natbib}

\appendix

\section{Human Annotation of Depth}
We hire two annotators who major in Linguistics and have received Bachelor degree to grade the depth of sampled triples. The triples are graded on a discrete scale of 1 to 4 where 1 represents the shallowest knowledge triple and 4 represents the deepest one.
For inter-annotator agreement, we calculate the Pearson correlation coefficient of the two annotators over the depth rank and the perplexity given by GPT-2 and Transformer-XL. The Pearson correlation over these two metrics given by two models are 0.67, 0.62, 0.58, and 0.61 respectively. 

\section{Experimental Settings for CKBC Task}
We adopt the same experimental settings for the results in Figure~4 and Table~1 in the main paper with ELMo, BERT and RoBERTa as backbones. 
Specifically, for experiments with BERT and RoBERTa, we employ BERT$_{large}$ and the RoBERTa$_{large}$ which have 24 layers, 16 self-attention heads and 1,024 hidden units. 
During training, the learning rate is set to 1e-5 with warmup over the first 8,000 steps and then decreases proportionally to the inverse square root of the number of steps, and the dropout rate is set to 0.1. For the ELMo model, we use the original two-layer bi-directional LSTM with 4,096 hidden units and obtain the final 512-dim contextualized representation of each word. All models are trained with 4,096 tokens per GPU and optimized with the Adam optimizer.

\section{Construction of the DCSK Dataset}
\label{sec:dataset_est}
We first use the pre-trained language models to generate candidate commonsense knowledge triples. However, such a method can only obtain limited knowledge triples. To further increase the quantity and the diversity of the candidate knowledge triples, we design simple strategies to propagate them to form new triple candidates. 
The propagated candidate triples are not directly derived from pre-trained language models and thus have a high chance to be deep commonsense triples. Finally, we use the depth rank given by GPT-2 introduced in the main paper to filter the deep commonsense triples from these candidates.

\subsection{Retrieving triples from Pre-trained LMs}
\label{subsec:retrieve_from_lms}
To encourage the language model to generate commonsense knowledge triples, we adapt language model to knowledge generation by fine-tuning the pre-trained language model on knowledge triples. The adapted task is formulated as follows: given a tuple $(h, r)$ representing head term $h$ and relation $r$ as input, our goal is to generate an appropriate tail term $t$ by a language model. 

We employ GPT-2 as our language model due to its impressive performance in text generation tasks. We denote the tokens in head term as $X^h = \{x_0^h, \dots, x_{|h|}^h\}$, the tokens in tail term as $X^t =\{x_0^t, \dots, x_{|t|}^t\}$, and the tokens that make up the relation as $X^r = \{x_0^r, \dots, x_{|r|}^r\}$. $|x|$ denotes the number of tokens of $x$. For a triple $(h, r, t)$, the corresponding input $X$ of the model is the concatenated sequence of the tokens in the triple:
\begin{equation}
    X = X^h \oplus X^r \oplus X^t
\end{equation}
where $\oplus$ denotes the concatenation operation. The sum of the word embeddings and the position embeddings of tokens in $X$ results in the initial representation $h_0$ of input.
GPT-2 stacks $n$ identical transformer blocks and applies the following transformations to encode hidden representations:
\begin{equation}
    h_l = {\rm transformer\_block} (h_{l-1}), \forall i\in [1, n]
\end{equation}

During training, we train the model to minimize the negative log-likelihood derived from the tail tokens:
\begin{equation}
    \mathcal{L} = - \sum_{l=|h| + |r|} ^{|h| + |r| + |t|} \log P(x_l | x_{<l})
\end{equation}
where $x_{<l}$ denotes the preceding tokens of the $l$-th token $x_l$. During inference stage, the model is supposed to auto-regressively generate $X^t$.

Considering the fact that there are multiple reasonable tail terms, we employ beam search in inference phase to generate multiple $t'$ for a $(h, r)$ pair. Then we obtain a triple set $\mathcal{S}_1$ consisting of commonsense knowledge triples $(h, r, t')$. 

\subsection{Knowledge Triple Propagation}
\begin{figure}[t]
\begin{center}
\includegraphics[width=0.95\linewidth]{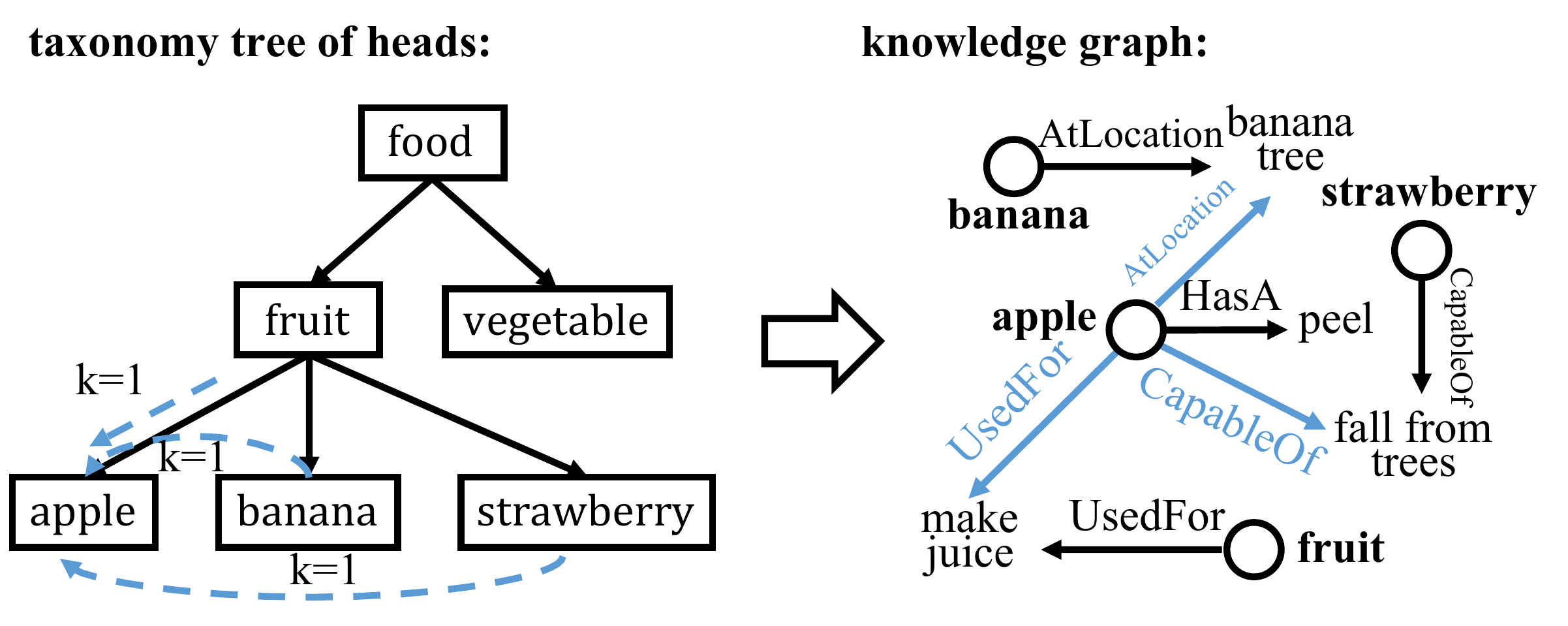}
\end{center}
\caption{The overview of our knowledge propagation based on the taxonomy tree. The blue dotted line represents the propagation path. $k$ denotes the propagation distance. The knowledge graph contains generated (black) and expanded (blue) relations.}
\label{fig:expand_rule}
\end{figure}

\label{subsec:know_propa}
Due to the limited size of $\mathcal{S}_1$, we further use simple strategies to propagate these triples to form new candidate triples. Motivated by the observation that human can learn new knowledge by analogy, we develop simple yet effective strategies to perform triple propagation.

Specifically, we define the hypernym-hyponym relation of the head terms with the help of WordNet. We regard the hypernym as the parent node of the corresponding hyponym head term, and the head terms sharing the same hypernym are regarded as the sibling nodes. We consider the relation and tail pair $(r, t)$ in a triple $(h, r, t)$ as the attribute of the head $h$. Knowledge propagation can be treated as transferring attributes of a node to its neighboring nodes according to the hypernym-hyponym relation. In more detail, we develop two propagation paradigms:
\begin{itemize}
    \item \textbf{horizontal propagation} Intuitively, sibling nodes may share the same attributes. For instance, if head term \textit{apple} and \textit{orange} have the same hypernym and are sibling nodes, the attribute (\emph{AtLocation}, tree) of \emph{apple} can be propagated to \emph{orange} to form a new candidate triple (orange, \textit{AtLocation}, tree).
    \item \textbf{vertical propagation} Hypernym terms usually have more general semantic meanings than hyponym terms, thus the attribute owned by the parent node can be propagated to its child nodes. For example, head term \textit{fruit} is the parent node of \textit{apple} in the taxonomy tree, then the attribute (\emph{AtLocation}, tree) can be propagated to form a new triple (apple, \emph{AtLocation}, tree).
\end{itemize}
An overview of horizontal and vertical propagation is illustrated in Figure~\ref{fig:expand_rule}.

After applying the above strategies to the triple set $\mathcal{S}_1$, we obtain a large amount of triple candidates which is denoted as $\mathcal{S}_2$. Since all these triple candidates are not directly derived from language model, which means these triples are inclined to be not consistent with the expression of language, we speculate that they are likely to be deep commonsense triples.

\subsection{Selection of Deep Commonsense Triples}
Given the triple candidate set $\mathcal{S}_2$, we remove the triples contained in the $\mathcal{S}_1$ and keep the left triples as the candidates of deep commonsense triples. 
We then use the depth rank introduced in the main paper to filter the candidates and obtain the deep commonsense triple candidates. 
We further hire two annotators majoring in Linguistics to check the validity of each triple. 
We discard the triples with conflict labels given by two annotators and only keep the triples that have the same human annotated labels. Finally, we select 2,000 examples for development set and test set respectively to form our DCSK evaluation dataset. 

\section{Effect of Context Pair Number}
\begin{figure}[t]
\begin{center}
\includegraphics[width=0.9\linewidth]{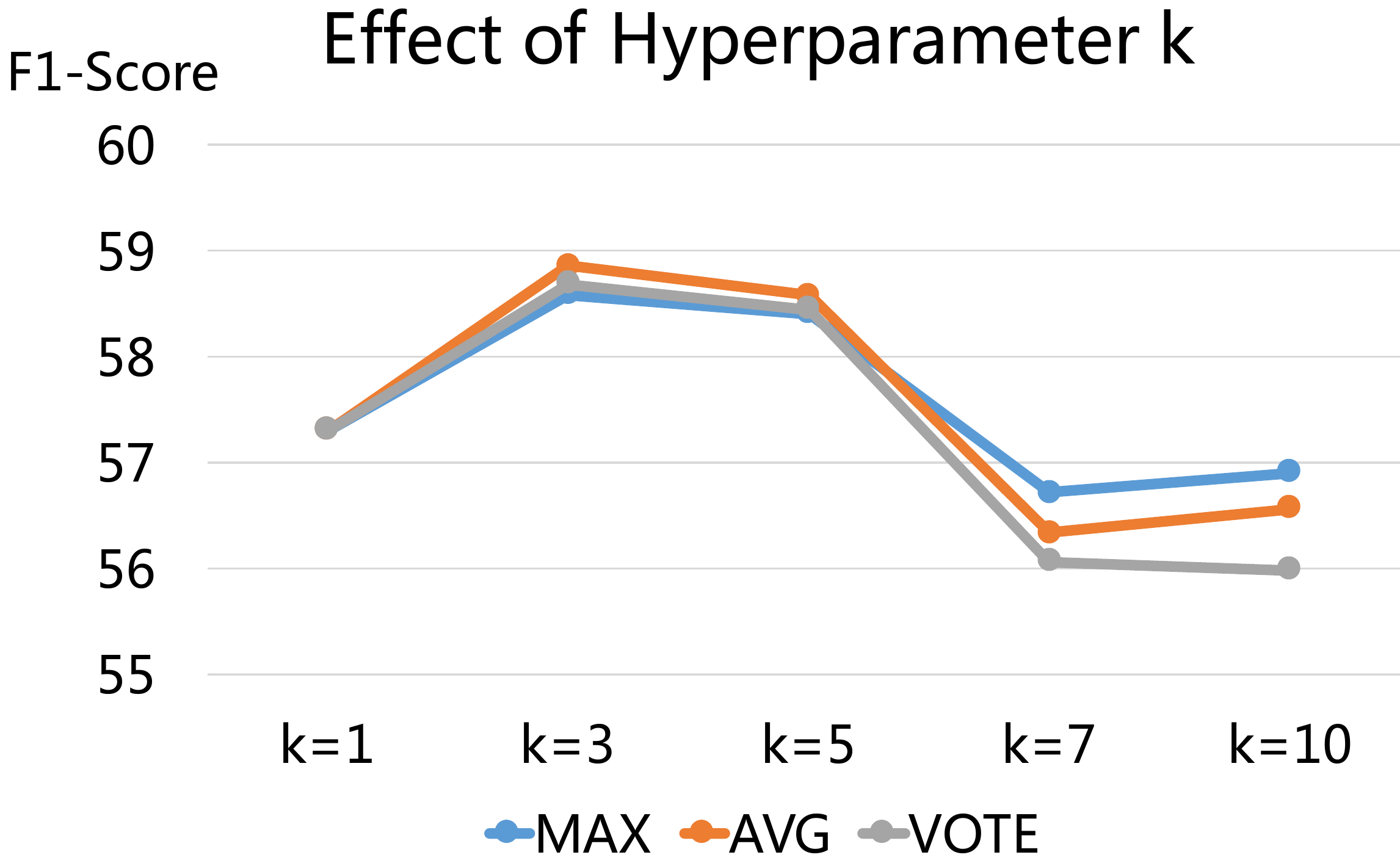}
\end{center}
\caption{Performance under different settings of hyper-parameter $k$ which represents the context pair number.}
\label{fig_para_k}
\end{figure}
Since our proposed method takes advantages of different context pairs, here we also explore the effect of context pair number $K$. We show the performance under different settings of $K$ in Figure~\ref{fig_para_k}. It first shows an increasing trend and then the performance declines. Although we select the sentence pairs with the most overlap words as input contexts, with only one sentence pair for a triple, the prediction seems to have more occasional noise. Therefore, at the beginning the performance is not the highest. When $K$ increases to $3$, we find a consistent performance improvements with all three inference methods, which suggests that including more context pairs is beneficial in making more accurate predictions. It is expected because more context pairs weaken the influence of occasional noisy pairs and make the final prediction based on different pairs more reliable than that based on single pair. However, as the $K$ further increases, we observe a performance drop with these inference methods. We hypothesize that more context pairs introduce more irrelevant and noisy information for prediction and lead to the performance drop. In practice, $K$ is recommended to set to 3, 4 or 5.

\section{Comparison of Inference Methods}

\begin{table}[t]
\begin{center}
\begin{tabular}{c|c|c|c}

\toprule 
    Methods     & Precision  & Recall & F1-score \\
\midrule

      Avg-Pred &51.40 & 68.09 &58.58 \\
      Max-Pred &51.25 &67.95 &58.48 \\
      Vote-Pred &50.97 &66.85 &55.98 \\

\bottomrule
\end{tabular}
\caption{Comparison of different inference methods on the DCSK dataset.}
\label{tab:vote}
\end{center}
\end{table}

We compare different inference methods on the DCSK dataset and demonstrate the results in Table~\ref{tab:vote}. These inference methods correspond to the sentence pair number $K=3$ since we empirically find it works best on the development set. We take F1-score as the main metric since it reflects a comprehensive result for recognizing valid entries. Avg-Prediction and Max-Prediction achieve comparable results. Max-Prediction takes the most confident prediction as the final result and Avg-Prediction method represents the average level of all prediction results. Both inference methods are relatively reliable. 
Although Vote-Prediction also takes all context pairs into account, the useful contexts and the noisy contexts have equal importance and thus this strategy is less robust than the other two methods. 

\end{document}